\documentclass[sigconf,nonacm]{acmart}

\AtBeginDocument{%
  }

\setcopyright{acmlicensed}
\copyrightyear{2026}
\acmYear{2026}
\acmDOI{XXXXXXX.XXXXXXX}
\acmConference[KDD '26]{32nd ACM SIGKDD Conference on Knowledge Discovery and Data Mining}{August 9--13,
  2026}{Jeju, Korea}
\acmISBN{978-1-4503-XXXX-X/2018/06}

\usepackage{subcaption}
\usepackage{booktabs}
\usepackage{siunitx}
\usepackage{graphicx}
\usepackage{pgfplots}
\pgfplotsset{compat=1.18}
\usepackage{xcolor}

\begin{document}

\title{PinCLIP: Large-scale Foundational Multimodal Representation at Pinterest}

\author{Josh Beal}
\email{jbeal@pinterest.com}
\affiliation{%
  \institution{Pinterest Inc.}
  \city{San Francisco}
  \state{CA}
  \country{USA}
}

\author{Eric Kim}
\email{erickim@pinterest.com}
\affiliation{%
  \institution{Pinterest Inc.}
  \city{San Francisco}
  \state{CA}
  \country{USA}
}

\author{Jinfeng Rao}
\email{marquisrao@pinterest.com}
\affiliation{%
  \institution{Pinterest Inc.}
  \city{San Francisco}
  \state{CA}
  \country{USA}
}

\author{Rex Wu}
\email{rexwu@pinterest.com}
\affiliation{%
  \institution{Pinterest Inc.}
  \city{San Francisco}
  \state{CA}
  \country{USA}
}

\author{Dmitry Kislyuk}
\email{dkislyuk@pinterest.com}
\affiliation{%
  \institution{Pinterest Inc.}
  \city{San Francisco}
  \state{CA}
  \country{USA}
}

\author{Charles Rosenberg}
\email{crosenberg@pinterest.com}
\affiliation{%
  \institution{Pinterest Inc.}
  \city{San Francisco}
  \state{CA}
  \country{USA}
}

\renewcommand{\shortauthors}{Beal et al.}

\begin{abstract}
While multi-modal Visual Language Models (VLMs) have demonstrated significant success across various domains, the integration of VLMs  
into recommendation and retrieval systems remains a challenge, due to issues like training objective discrepancies and serving efficiency bottlenecks. This paper introduces PinCLIP, a large-scale visual representation learning approach developed to enhance retrieval and ranking models at Pinterest by leveraging VLMs to learn image-text alignment. We propose a novel hybrid Vision Transformer architecture that utilizes a VLM backbone and a hybrid fusion mechanism to capture multi-modality content representation at varying granularities. Beyond standard image-to-text alignment objectives, we introduce a neighbor alignment objective to model the cross-fusion of multi-modal representations within the Pinterest Pin-Board graph. Offline evaluations show that PinCLIP outperforms state-of-the-art baselines, such as Qwen~\cite{li2026qwen3}, by 20\% in multi-modal retrieval tasks. 
Online A/B testing demonstrates significant business impact, including substantial engagement gains across all major surfaces in Pinterest.
Notably, PinCLIP significantly addresses the "cold-start" problem, enhancing fresh content distribution with a 15\% Repin increase in organic content and 8.7\% higher click for new Ads.


\end{abstract}




\newcommand{\jb}[1]{{\color{red}josh: #1}}
\newcommand{\ek}[1]{{\color{cyan}eric: #1}}
\newcommand{\jr}[1]{{\color{purple}jinfeng: #1}}
\newcommand{\rw}[1]{{\color{orange}rex: #1}}
\newcommand{\dk}[1]{{\color{green}dmitry: #1}}
\newcommand{\chr}[1]{{\color{blue}chuck: #1}}

\maketitle

\section{Introduction}
Large Language Models (LLMs) have achieved great success in recent years on revolutionizing a vast number of domains starting with the introduction of GPT~\cite{radford2018improving}. The introduction of the Transformer architecture~\cite{vaswani2017attention} provides a more effective way to handle long-range dependencies in text, which led to the development of models like BERT~\cite{devlin2019bert}, T5~\cite{raffel2020exploring}, and GPT~\cite{radford2018improving} that pioneered the use of large-scale, unsupervised pre-training to generate coherent and contextually relevant text. Subsequent breakthroughs, such as instruction tuning~\cite{chung2024scaling} and reinforcement learning from human feedback (RLHF)~\cite{ouyang2022training}, have further refined these models, enabling them to follow complex instructions and produce safer, more helpful responses.

Inspired by the generative design of LLMs, early research, such as SASRec~\cite{kang2018self} or BERT4Rec~\cite{sun2019bert4rec}, brings generative modeling with unidirectional or bidirectional attention through transformers~\cite{vaswani2017attention} to sequential recommendation, though not directly leveraging the pretrained LLMs for content representations. Subsequent research, such as ZesRec~\cite{ding2021zero}, UniSRec~\cite{hou2022towards}, and others~\cite{li2023text,ren2024representation,hou2024bridging}, advanced this by leveraging pre-trained LLMs to encode textual descriptions of items and user behaviors. A separate line of work~\cite{penha2020does,yang2021improving,he2023large} has explored using prompt-tuning~\cite{lester2021power} to directly elicit recommendations from LLMs based on user history or conversations. More recently, a trend has emerged toward pre-training large, GPT-style generative models on user interaction sequences for representation learning in recommender systems~\cite{zhai2024actions,chen2025pinfm,wang2025scaling,yi2025recgpt}.

Despite this progress, existing approaches have notable limitations. First, many models adopt LLM architectures or pre-training strategies~\cite{kang2018self,sun2019bert4rec,zhou2020s3,zhai2024actions,chen2025pinfm,wang2025scaling} without directly harnessing the rich semantic representations from pre-trained foundation models. Second, the primary focus has remained on textual data~\cite{ding2021zero,hou2022towards,li2023text,ren2024representation,hou2024bridging}, with other modalities such as images and video being largely underexplored. Finally, most studies are conducted on small-scale academic datasets, providing insufficient insight into the challenges of deploying these models scalably in real-world production environments.

Overall, we believe that there are several fundamental challenges that impede the integration of LLMs into production recommender systems (RecSys). Primarily, the optimization objectives differ between the two paradigms. Modern LLMs like GPT~\cite{radford2018improving} are generative models trained to predict the next token, whereas RecSys models are typically trained for discriminative tasks -- such as predicting user engagement with a video -- utilizing cross-entropy loss functions. Secondly, the operational requirements of RecSys necessitate high throughput and low latency to serve live traffic. The substantial model sizes of LLMs, which can contain billions of parameters, pose a practical challenge for deployment in production RecSys due to their considerable computational cost and inherent latency. Finally, LLMs and large vision models (VLMs) are trained on multi-modal data, where inputs are tokenized into granular units like text tokens or image patches. In contrast, RecSys models leverage manually engineered features and learn from sparse implicit feedback derived from a potentially vast corpus of items. These critical architectural and data-centric differences present substantial technical hurdles for the seamless integration of LLMs into RecSys.

This paper introduces \textbf{PinCLIP}, a large-scale visual representation learning approach developed to enhance retrieval and ranking models at Pinterest. We develop a novel hybrid Vision Transformer architecture, leveraging a VLM backbone and a multi-level fusion process to characterize diverse content granularities. To enrich the representation space, we supplement standard image-to-text alignment objectives with a neighbor alignment objective. This allows the model to capture the complex interplay of multi-modal signals within the structural context of the Pinterest Pin-Board graph. We also leveraged Matryoshka Representation Learning (MRL)~\cite{kusupati2022matryoshka} to reduce the dimension of learned PinCLIP representation for more efficient downstream adoption. To scale the training and inference of PinCLIP, we employed multiple optimization techniques, including Flash Attention~\cite{dao2022flashattention, shah2024flashattention}, funneling operations~\cite{dai2020funnel, choi2025revisiting}, and activation checkpointing to enable multi-node training on billions of images. 
The resulting representations are integrated into dozens of machine learning production models, significantly enhancing user engagement across different surfaces in Pinterest.


The key contributions of this work are as follows:
\begin{itemize}
\item We introduce a novel hybrid vision transformer architecture that leverages pretrained VLMs to learn multi-modal representations from a massive, real-world product with hundreds of millions of users.
\item We present a practical recipe on how to build an effective multi-modality content signal on a large-scale image dataset, including our best practices on data selection, model architecture, learning objectives, efficiency enhancements and full productionization.
\item Our proposed visual representations demonstrate over 20\% offline metric gains against strong baselines like Qwen~\cite{li2026qwen3}. 
surfaces in Pinterest, including Homefeed, Search and Related Pins.  
More importantly, online A/B testing demonstrates significant business impact, including substantial engagement gains across all major
surfaces in Pinterest, including Homefeed, Search and Related Pins. 
\item We also show that our learned multi-modality content representations are highly effective at promoting fresh content on the platform with up to 15\% more Repins on organic content and 8.7\% more clicks on new Ads, which is a crucial step toward solving the longstanding cold-start problem in recommender systems.
\end{itemize}

\section{Related Work}
\subsection{Visual Language Models}
The evolution of visual-language models (VLMs) began with a foundational shift from CNNs to powerful Transformer-based vision backbones like the Vision Transformer (ViT)~\cite{dosovitskiy2020image}, which treated image patches as sequential tokens, and the more efficient Swin Transformer~\cite{liu2021swin}, which introduced a hierarchical, windowed self-attention mechanism. These architectures paved the way for large-scale pre-training strategies, exemplified by OpenAI's CLIP~\cite{radford2021learning}, which pioneered contrastive learning on massive web-scale image-text pairs to achieve remarkable zero-shot generalization. Subsequent models like BLIP~\cite{li2023blip} refined this approach by introducing image-to-text generation tasks, and SigLIP~\cite{zhai2023sigmoid} simplified contrastive learning with sigmoid loss to achieve substantial efficiency gains. The current frontier is defined by scaling, with public models such as InternVL~\cite{chen2024expanding, zhu2025internvl3} and Alibaba's Qwen-VL~\cite{bai2023qwen, li2026qwen3}, and closed models like Gemini~\cite{comanici2025gemini}, pushing parameter counts into the billions. These state-of-the-art systems demonstrate that increasing model and data scale continues to yield significant gains, enabling more sophisticated and fine-grained multimodal reasoning capabilities.

\subsection{LLMs for Recommender Systems}

Drawing inspiration from the generative capabilities of LLMs, initial studies in sequential recommendation adopted Transformer architectures. Models like SASRec~\cite{kang2018self}, BERT4Rec~\cite{sun2019bert4rec}, and $S^3$-Rec~\cite{zhou2020s3} utilized generative modeling with either unidirectional or bidirectional attention, although they did not use pre-trained LLMs for content understanding. A later wave of research, including ZesRec~\cite{ding2021zero}, UniSRec~\cite{hou2022towards} and others~\cite{li2023text,ren2024representation,hou2024bridging}, advanced the field by employing pre-trained LLMs to encode the textual features of items and user interactions. Concurrently, another research direction explored prompt-tuning LLMs to directly generate recommendations from user histories or dialogues, as seen in works by \cite{penha2020does,yang2021improving,he2023large}. More recently, the field has shifted towards pre-training large, GPT-style generative models directly on sequences of user behavior to learn powerful representations for recommendation tasks~\cite{zhai2024actions,chen2025pinfm,wang2025scaling,yi2025recgpt}. Fu et al.~\cite{fu2024iisan} utilize multi-modality models as item encoders for sequential recommendation and introduce a decoupled PEFT mechanism to boost efficiency. Their approach still requires multi-modality model inferences for each item in the user sequence, which could lead to significant scalability challenges when scaling to a production environment with hundreds of millions of users and items. 

A concurrent work from Giahi et al.~\cite{giahi2025vl} also learns multi-modal embeddings using a CLIP-style objective by aligning object-level image crops with LLM-generated text summaries. Our approach differs in several critical aspects. First, unlike their method which models vision and text through separate encoders, our model employs a cross-modal fusion architecture to learn a richer, unified semantic representation. Second, we introduce a novel neighbor alignment task that is critical for learning effective representations. Consequently, our approach produces a holistic image embedding not explicitly defined in their object-centric method, demonstrating strong performance on both retrieval and ranking tasks, whereas their evaluation is focused primarily on retrieval.

\section{Method}

\begin{figure*}[t]
 \centering 
 \includegraphics[width=\textwidth]{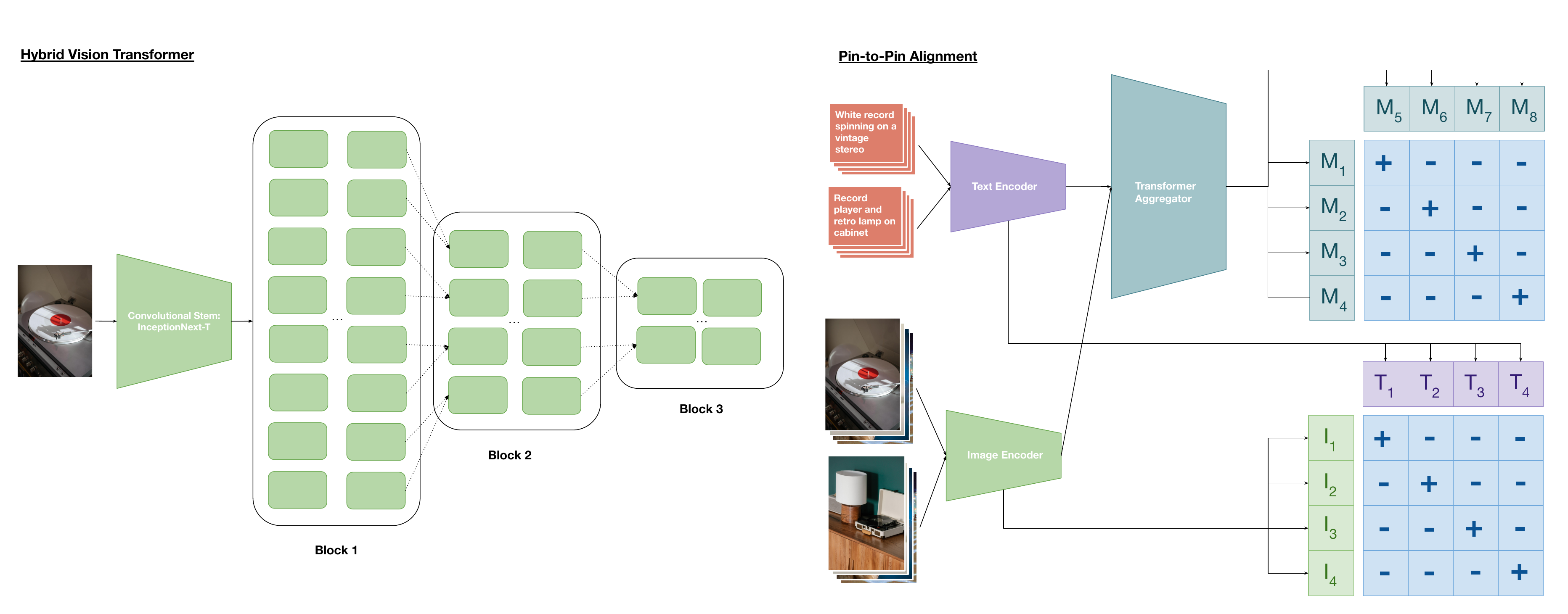}
  \caption{Overview of the PinCLIP fusion model architecture, highlighting the design of the Hybrid Vision Transformer backbone and the multimodal contrastive objectives for jointly learning image-text alignment and Pin-to-Pin alignment.}
 \label{fig:model-comparison}
\end{figure*}

Pinterest, as a visual-centric platform, is primarily used by its users to discover visual inspiration across a wide spectrum of topics. To enhance the functionality of the platform, our objective is to construct a general and effective multimodal representation learning system. This system is designed to benefit numerous downstream machine learning models in production, spanning various surface areas such as Homefeed, Related Pins, Search, and Ads. To achieve this goal, we develop a novel hybrid Vision Transformer architecture, leveraging a VLM backbone and a multi-level fusion process to learn the multi-modality content representation at various level of granularities. Our model architecture is detailed in the following section.

\vspace{-1em}
\subsection{Model Architecture}
This model architecture includes an image encoder $E_\mathsf{img}(\cdot)$ and a text encoder $E_\mathsf{txt}(\cdot)$, where the same text encoder is used to process different text sources. The fusion encoder $E_\mathsf{fsn}(\cdot)$ computes the image and text representations of the Pin. Given a Pin $j$ with image $I_j$ and an associated text field $T_j$ (e.g., title/description), the image encoder $E_\mathsf{img}$ produces a sequence of visual tokens and the text encoder $E_\mathsf{txt}$ produces a sequence of textual tokens:
\begin{align}
\label{eq:encoders}
\tilde V_j &= E_\mathsf{img}(I_j) \in \mathbb{R}^{N_v \times d},\\
\tilde S_j &= E_\mathsf{txt}(T_j) \in \mathbb{R}^{N_t \times d},
\end{align}
where $N_v$ is the number of visual tokens and $N_t$ is the number of textual tokens.

\subsubsection{Multi-Modality Encoder}
We consider two in-house pretrained image encoders~\cite{beal2022billion} of varying sizes, which significantly outperform leading open-source vision models for this application. These in-house models consist of a convolutional neural network stem and a Funnel Vision Transformer trunk~\cite{dai2020funnel, dosovitskiy2020image}. Prior work has shown that early convolutions accelerate and stabilize training, while improving classification accuracy~\cite{xiao2021early}. We find that the usage of funneling operations can significantly reduce the computational cost and memory usage of pretraining while maintaining strong retrieval quality. We pretrain the image encoder using the billion-scale, weakly-supervised classification task that was proposed in our prior work~\cite{beal2022billion}. We additionally consider two open-source text encoders of varying sizes, specifically the SigLIP multilingual text encoders ~\cite{zhai2023sigmoid}, which outperformed other pretrained text encoders in end-to-end training.


\subsubsection{Multi-Modality Fusion}
\label{sec:fusion}

The fusion encoder $E_\mathsf{fsn}(\cdot)$ computes the image and text representations of the Pin, and applies a Transformer aggregator, consisting of 2 layers, to the unpooled visual and textual tokens, yielding a multimodal embedding $M'_j = E_\mathsf{fsn}(I_j, T_j)$. For this task, we use the descriptive text associated with the Pin. 

To enable deep modality integration, we fuse unpooled visual and textual tokens with a small Transformer aggregator of $L_f$ layers. We first concatenate modality token sequences:
\begin{align}
Z_j^{(0)} = [\tilde V_j \, \| \, \tilde S_j] \in \mathbb{R}^{N \times d}, \quad N = N_v + N_t.
\end{align}
We then apply $L_f$ standard Transformer layers with vanilla self-attention:
\begin{align}
Z_j^{(\ell+1)} = \mathrm{TransformerLayer}\!\left(Z_j^{(\ell)}\right), \quad \ell=0,\dots,L_f-1.
\end{align}

Finally, we compress the multimodal token sequence into a single embedding $r_j \in \mathbb{R}^{d}$ using a single multi-head attention pooler. Let $q \in \mathbb{R}^{1\times d}$ be a learned pooling query; then
\begin{align}
r_j = \mathrm{MHA}(q, Z_j^{(L_f)}, Z_j^{(L_f)}) \in \mathbb{R}^{1\times d}.
\end{align}
We denote the fused Pin embedding by $M'_j$ (and use $\ell_2$ normalization before matching) where:
\begin{align}
M'_j = \frac{r_j}{\|r_j\|_2}.
\end{align}


\subsubsection{Objectives} 
We consider using multiple contrastive loss terms to capture multi-modality feature alignment in varying granularity. \\

\noindent \textbf{Image-to-text Alignment}: We start with image-to-text alignment that captures interactions between the query image and different text source signals. We consider two text signals: 1) the descriptive text; 2) the keyword text. The descriptive text consists of the Pin title, the detailed Pin description, or the synthetic caption (generated by an open-source VLM~\cite{li2022blip, liu2024improved}), selected in order of availability. Likewise, the keyword text consists of the engaged search keywords or the extracted Pin keywords, selected in order of availability. The descriptive text is usually much longer than the keyword text. We found that it was more effective to coalesce the text sources as described above, rather than concatenate all of the available text into a single text string. Several works have observed that CLIP text encoders exhibit relatively short effective context length~\cite{zhang2024long}, and highly detailed image descriptions can negatively affect contrastive learning performance~\cite{liu2024clips}. 


We adopt the sigmoid loss from SigLIP, which is a simple pairwise loss that outperforms the traditional softmax loss used in previous work~\cite{radford2021learning}. This objective yields better downstream performance, especially at smaller batch sizes, while also improving training efficiency. By processing each image-text pair independently, it avoids the heavy communication between nodes that is required to compute the global normalization factor. We describe the  efficient ``chunked'' implementation below, which further reduces the significant communication costs of multi-node training.

Consider a mini-batch $\mathcal{B} = ((I_1, T_1), \dots, (I_n, T_n))$ of image-text pairs, and let $\mathcal{B}' = ((I'_1, T'_1), \dots, (I'_n, T'_n))$ be the encoded representations where $V'_j = E_\mathsf{img}(I_j)$ and $S'_j = E_\mathsf{txt}(T_j)$ for $j \in \{ 1, \dots, n \}$, which are the image and text representations from Equation~(\ref{eq:encoders}) and (2). For a set of devices of size $D$, we let $b = \frac{|\mathcal{B}|}{D}$ be the per-device batch size. We define the per-instance loss as:

\begin{equation}
\mathcal{L}_{ij} = \log (1 + e^{\mathbf{z}_{ij}(-t\mathbf{x}_i \cdot \mathbf{y}_j+c)})
\end{equation}

\noindent where $\mathbf{x}_i = \frac{V'_i}{\lVert V'_i \rVert_2}$ and $\mathbf{y}_j = \frac{S'_j}{\lVert S'_j \rVert_2}$ are the normalized image/text representations, and $\mathbf{z}_{ij}$ is the label for a given image-text pair, equaling 1 if they originate from the same Pin and $-1$ otherwise, $t$ is the temperature, and $c$ is the bias. To improve convergence, we set $t=\log 10$ and $c=-10$ at initialization.

We define the image-to-text alignment loss as follows:
\begin{equation}
\mathcal{L_\texttt{I2T}} = -\frac{1}{\lvert \mathcal{B} \rvert}
\sum_{d_i=1}^{D}
\sum_{d_j=1}^{D}
\sum_{i=b d_i}^{b(d_i+1)}
\sum_{j=b d_j}^{b(d_j+1)}
\mathcal{L}_{ij}.
\end{equation}


\noindent \textbf{Pin-to-Pin Alignment}: 
Besides the image-to-text alignment, we introduce a new objective that aligns the representations of Pin pairs (i.e., between query Pins and target Pins). These Pin pairs are sampled from a large-scale (pruned) Pin-Board graph, where an edge exists between a Pin and a Board if a user has saved the Pin to a given Board, which is a collection of Pins on the same topic. This manual curation signal yields an extremely valuable bipartite graph, which has been widely used at Pinterest for efficient candidate generators~\cite{eksombatchai2018pixie} and engagement-based embeddings~\cite{eksombatchai2018pixie, ying2018graph, badrinath2025omnisage}.

We adapt the sigmoid loss to enforce the similarity of the paired multimodal representations. As above, we normalize the resulting multimodal embedding. We let $\mathbf{u}_i = \frac{M'_i}{\lVert M'_i \rVert_2}$ and $\mathbf{v}_j = \frac{M'_{j+n/2}}{\lVert M'_{j+n/2} \rVert_2}$ are the normalized multimodal representations for $i,j \in [\frac{n}{2}]$ and $\mathbf{w}_{ij}$ is the label for a given  pair of Pins, equaling 1 for neighbor Pins and $-1$ otherwise, $t$ is the temperature, and $c$ is the bias. We extend the pairwise loss from~\cite{zhai2023sigmoid} and define the Pin-to-Pin alignment loss as:

\vspace{-0.5em}

\begin{align}
\label{eq:fusion_loss}
\mathcal{L}_{ij} &= \log (1 + e^{\mathbf{w}_{ij}(-t\mathbf{u}_i \cdot \mathbf{v}_j+c)}) \\
\mathcal{L}_\texttt{P2P} &= \sum_{i}^D \sum_{j}^D \mathcal{L}_{ij} 
\end{align}
\vspace{-0.5em}

The total loss is a sum of the image-to-text and Pin-to-Pin alignment loss:
\begin{equation}
\label{eq:total_loss}
\mathcal{L} = \mathcal{L}_\texttt{I2T} + \mathcal{L}_\texttt{P2P} \\
\end{equation}

\subsection{Data Curation}

\begin{figure}[t]
    \centering
    \includegraphics[width=0.45\textwidth]{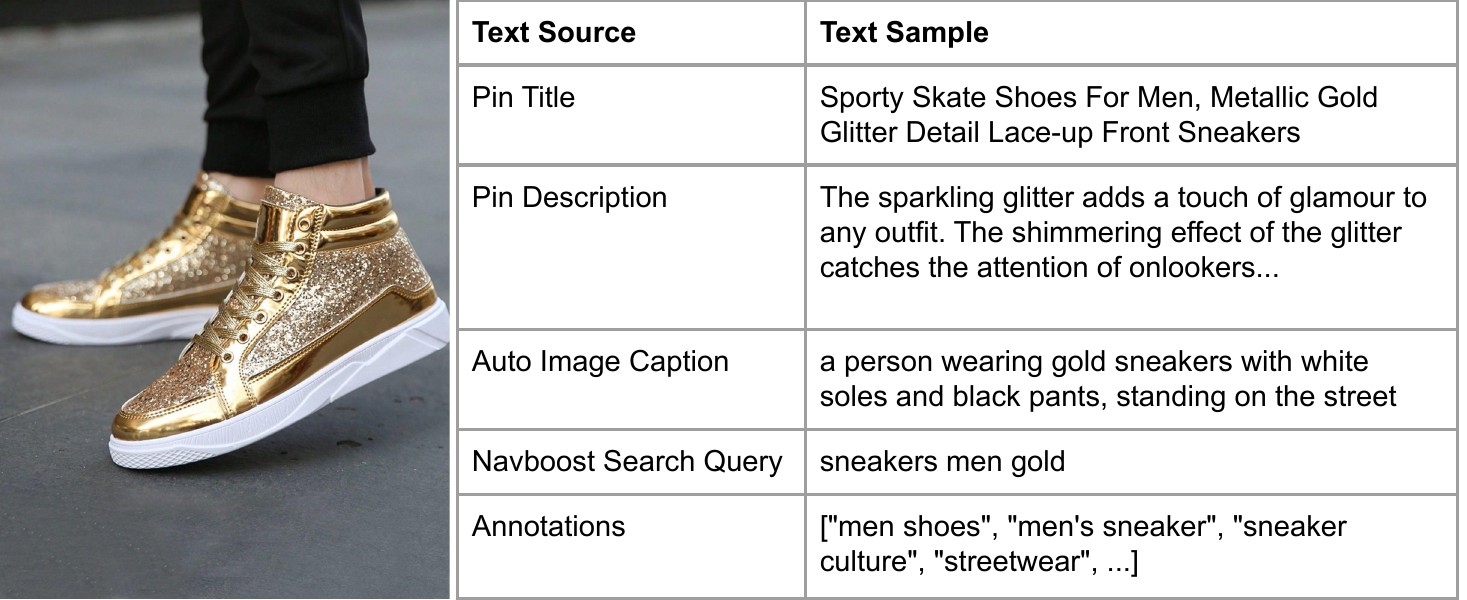}
    \caption{
Illustration of the text-image dataset.
Each image (``Pin'') is associated with multiple text signals.
We use longer-form \emph{descriptive} text signals (title, description, image caption) and shorter-form \emph{keyword} text signals (navboost search query, annotations).}
    \label{fig:pin_image_and_text}
\end{figure}

On Pinterest, each Pin is associated with an image and title, along with an optional text (known as description) and link. We curate a large dataset from our platform to enable large-scale multi-modality representation learning. The dataset contains about $890$M unique images (aka. Pins) with high quality text sources for training PinCLIP. Since there is a significant
volume of Pins on Pinterest (about 30\%) lack associated titles or descriptions, or have noisy and/or irrelevant title or description, we consider multiple ways to augment the quality of our text signals. 
See Figure~\ref{fig:pin_image_and_text} for an illustration of the text signals we use.
Specifically, we consider the following text sources.
\begin{itemize}
    \item Pin title: the native title associated with an image.
    \item Pin description: the native description of an image.
    \item Image caption: we employ an off-the-shelf image captioning model (BLIP~\cite{li2023blip} and LLaVA~\cite{liu2023visualinstructiontuning}) to generate a synthetic description of the image. 
    \item Navboost query: On Pinterest, users can search different topics and save their liked Pins on the Search feed. We link the saved Pins with the search queries, and use the search queries as keywords to enhance the textual representation of Pins.
    \item Annotation: Each Pin is associated with multiple short-text annotations. These annotations are produced based on n-grams of user queries and Pin text sources, along with a learned ranking score.
\end{itemize}

To further improve the data quality, we calculated image-text alignment scores (via the CLIPScore~\cite{radford2021learning} method using OpenAI's CLIP-L as the reference model) and filtered out all pairs whose alignment score was below a tuned threshold.

\subsubsection{Neighbor Pairs} To overcome the limitations of contrastive learning with the basic image-to-text alignment objective, which only focuses on alignment over an individual Pin, we construct a dataset that contains pairs of Pins that are engaged by the same set of users. For each query Pin, we sample $N=5$ positive Pins from the logs to produce $N$ query-positive pairs. The intent is that each query Pin and positive Pin should be semantically similar to each other. The sampling is done by performing random walks over the Pin-Board graph, starting at the query node, aggregating the visited nodes, and choosing the top $N$ visited nodes. For context, the Pin-Board graph is a graph data structure where each node is a Pin or a Board, and an (undirected) edge connects a Pin to a Board if a Pin is present in a Board. For efficiency, we first compute the top $K=50$ pairs for each query offline, and then sample $N$ pairs from the cached results. We observed a small benefit to weighted sampling by the visit count vs. uniform random sampling. In total, we sampled around 2.4B Pin pairs for training PinCLIP. It's worth noting that the above neighbor pair dataset can also be sampled from standard user engagement data, such as from Pin pairs that are engaged by the same user in one session.

\subsection{Training Efficiency}
The vision encoder design is a critical factor in improving the training efficiency. Adopting a simple Vision Transformer backbone dramatically increases the GPU resource requirements (by a factor of 2-4x) to achieve the same level of performance. Furthermore, we observed benefits from partial freezing of the vision encoder. This approach resulted in no degradation of the model quality in our experiments. In fact, freezing several Transformer layers was found to improve the downstream performance, similar to prior findings on locked image-tuning (LiT)~\cite{zhai2022lit}. We attribute this result to the high quality of the pretrained in-house image encoder, which is trained on a multi-label classification task~\cite{beal2022billion}. To further improve the training efficiency, we adopted FlashAttention-2~\cite{dao2023flashattention} and activation checkpointing techniques, which improved training throughput by 15\% and reduced GPU memory usage by 32\%.

\subsection{Signal Productionization}


\textbf{Matryoshka Representation Learning}: To reduce serving costs and ease signal adoption for downstream consumers, we explored reducing the embedding dimensionality via Matryoshka Representation Learning (MRL)~\cite{kusupati2022matryoshka}. MRL is a training paradigm that encodes information at multiple granularities within a single embedding vector, much like the nested structure of Russian Matryoshka dolls. By optimizing the loss across different prefix lengths of the same vector, MRL ensures that the first $k$ dimensions of an embedding are themselves a representative and performant latent space. This allows for adaptive retrieval; a system can use smaller, truncated embeddings for high-speed initial candidate generation (coarse-to-fine) and only utilize the full-dimensional vector when maximum precision is required, significantly reducing computational overhead and memory usage without requiring separate models for different scales. We train MRL with the following loss:

\begin{equation}
\mathcal{L}_{MRL}(x; \theta) = \sum_{k \in \mathcal{K}} c_k \cdot \mathcal{L}(f(x; \theta)_{1:k}; \mathbf{W}_k)
\end{equation}

\noindent where $f(x; \theta)_{1:k}$ represents the first $k$ dimensions of the output embedding, $W_k$ is a linear classifier/projection head specific to the $k$-dimensional representation, $c_k$ is the weight assigned to each scale, and $\mathcal{L}$ is the contrastive loss function.

We train the PinCLIP model with the MRL loss with the following prefix and weight settings: $(k, c_k)_i$ = [(64, 0.1), (128, 0.1), (256, 1.0)].
The MRL losses for all prefixes were added to the total loss in Equation~(\ref{eq:total_loss}). We publish the reduced-dimensionality embeddings for downstream consumers to ingest, since ingesting the full 256d embeddings could be infeasible for cost and resource reasons. We found that the 64d prefix embedding offers an acceptable tradeoff between representation quality and serving costs. To ensure that the embeddings can be easily utilized for downstream retrieval systems like approximate nearest neighbor search via inner-product distance, we perform L2 normalization of the prefix embedding.

\vspace{-0.5em}
\begin{align*}
& x_q = \text{round}\left(\frac{x}{s} + z\right)\\
& x_{dq} = (x_q - z) \cdot x
\end{align*}
\begingroup\vspace*{-\baselineskip}
\captionof{figure}{
Global affine embedding quantization.
$x$ is the original full-precision embedding.
$x_q$ is the quantized (lower-precision) embedding.
$x_{dq}$ is the reconstructed embedding used in downstream tasks.
Scalar $s, z$ are the scale and zero-point parameters respectively.
}
\label{fig:affine_quant}
\vspace*{\baselineskip}\endgroup
\vspace{-0.5em}

\noindent \textbf{Embedding Quantization}: To reduce serving costs, we implemented post-training quantization (PTQ) using int8 global affine quantization. 
Specifically, we transform the PinCLIP embedding vector from float16 into int8 representations via the quantization and dequantization transformations shown in Figure~\ref{fig:affine_quant}.
The $s, z$ scalar parameters are determined empirically to maximize reconstruction and retrieval recall on a representative sample of embeddings. We used $s=\frac{0.5}{127}$, and $z=0$. This approach compresses a 256d vector from 4096 bits to 2048 bits—reducing the total embedding size to 50\% of its original footprint and delivering a proportional reduction to serving costs. 
We verified that this affine quantization results in negligible impact on offline evaluation metrics.


\section{Experiments}

\subsection{Evaluation Setup}
We design our offline evaluation suite to reflect the real-world performance of the model in the production systems at Pinterest. We consider five key evaluation tasks: PinText image-text retrieval, Related Pins image-image retrieval, Related Pins multimodal retrieval, Search text-image retrieval and Search multimodal retrieval. For the multimodal retrieval tasks, we use the fusion embedding as the Pin representation, whereas the image embedding is used for the other surface evaluations.

We sample 30k pairs from the PinText training dataset to form the PinText evaluation dataset. For Related Pins and Search, we sample 80k engaged pairs, where a user saved the target Pin for a given text search query or query Pin. We include a 1.5M distractor set for the surface evaluations. For a proper evaluation, we ensure that the evaluation sets are disjoint from the training sets.

We compute the Recall@K metric for each evaluation task. Specifically, we evaluate whether the proposed method can retrieve the correct positive embedding among a set of random negative embeddings, given the query embedding. Given a set of query embeddings $Q$, positive embeddings $P$, and negative embeddings $N$, the Recall@K metric is defined below, where $|Q|$ denotes the number of query embeddings and the similarity metric is $s(n_1, n2) = n_1^T n_2$.
\[
\mathrm{Recall}_K(Q,P,N)
= \frac{1}{|Q|}
\sum_{i=1}^{|Q|}
\mathbf{1}\!\left\{
\left| \{\, n \in N \mid s(Q_i,n) \ge s(Q_i,P_i) \,\} \right| < K
\right\}
\]

\subsection{Training Procedure}
We train with a global batch size of 32,768 for the image-text alignment task and 4,096 for the neighbor alignment task using 64 H100 GPUs. We train for 150,000 steps, processing $\approx$6B images in total. We use a warmup phase of 6,000 steps, followed by cosine decay of the learning rate to zero. The base learning rate is $2 \times 10^{-4}$, and the optimizer is Lion~\cite{chen2023symbolic}. We found it beneficial to employ a 10x lower learning rate for the text encoder. We used a weight decay value of $5 \times 10^{-4}$ for the image encoder and fusion aggregator and a weight decay value of $5 \times 10^{-1}$ for the text encoder based on grid search.

\begin{figure}[t]
\centering
\captionsetup[subfigure]{justification=centering}
\setlength{\tabcolsep}{12pt}

\begin{subfigure}[t]{0.32\textwidth}
\centering
\caption{PinText}
\vspace{1mm}
{\small
\begin{tabular}{l *1{S}[table-format=2.1]}
\toprule {Method Name} & {R@1}\\
\midrule 
\multicolumn{2}{c}{Image-to-Text Retrieval} \\
\midrule 
CLIP-B~\cite{radford2021learning} & 29.4 \\
CLIP-L~\cite{radford2021learning} & 35.7 \\
mSigLIP-So~\cite{zhai2023sigmoid} & 58.6 \\
PEcore-G~\cite{bolya2025perception} & 60.5 \\
SigLIP2-g~\cite{tschannen2025siglip} & 62.0 \\
MetaCLIP2-g~\cite{nyandwi2025grounding} & 63.8 \\
Qwen3-VL-Embedding-8B~\cite{li2026qwen3} & 32.0 \\
\midrule 
PinCLIP Image-Text & \textbf{76.7} \\  
PinCLIP Fusion & \underline{76.4} \\
\bottomrule 
\end{tabular}
}
\label{tab:pintext_i2t}
\end{subfigure} \hfill

\begin{subfigure}[t]{0.32\textwidth}
\centering
\caption{Search}
\vspace{1mm}
{\small
\begin{tabular}{l *1{S}[table-format=2.1]}
\toprule {Method Name} & {R@10} \\
\midrule 
\multicolumn{2}{c}{Text-to-Image Retrieval} \\
\midrule 
CLIP-B~\cite{radford2021learning} & 9.3 \\
CLIP-L~\cite{radford2021learning} & 12.4 \\
mSigLIP-So~\cite{zhai2023sigmoid} & 24.3 \\
PEcore-G~\cite{bolya2025perception} & 23.6 \\
SigLIP2-g~\cite{tschannen2025siglip} & 26.7 \\
MetaCLIP2-g~\cite{nyandwi2025grounding} & 27.4 \\
\midrule 
PinCLIP Image-Text & 35.1 \\  
PinCLIP Fusion & \textbf{44.5} \\

\midrule 
\multicolumn{2}{c}{Multimodal Retrieval} \\
\midrule
OmniSearchSage~\cite{agarwal2024omnisearchsage} & 34.9 \\  
Qwen3-VL-Embedding-8B~\cite{li2026qwen3} & 26.2 \\
\midrule
PinCLIP Fusion & \textbf{47.1} \\
\bottomrule
\end{tabular}
}
\label{tab:search_t2i_and_mm}
\end{subfigure} \hfill

\begin{subfigure}[t]{0.32\textwidth}
\centering
\caption{Related Pins}
\vspace{1mm}
{\small
\begin{tabular}{l *1{S}[table-format=2.1]}
\toprule
\cmidrule(lr){1-2} {Method Name} & {R@10} \\
\midrule
\multicolumn{2}{c}{Image-to-Image Retrieval} \\
\midrule
CLIP-B~\cite{radford2021learning} & 19.7 \\
CLIP-L~\cite{radford2021learning} & 22.7 \\
mSigLIP-So~\cite{zhai2023sigmoid} & 35.1 \\
PEcore-G~\cite{bolya2025perception} & 35.3\\
SigLIP2-g~\cite{tschannen2025siglip} & 35.7 \\
MetaCLIP2-g~\cite{nyandwi2025grounding} & 33.8 \\
Unified Embedding~\cite{zhai2019learning} & 31.4 \\
\midrule 
PinCLIP Image-Text & 39.6 \\  
PinCLIP Fusion & \textbf{44.5} \\
\midrule 
\multicolumn{2}{c}{Multimodal Retrieval} \\
\cmidrule(lr){1-2}
Qwen3-VL-Embedding-8B~\cite{li2026qwen3} & 23.6 \\
\midrule
PinCLIP Fusion & \textbf{57.2} \\
\bottomrule
\end{tabular}
}
\label{tab:relatedpins_i2i_and_mm} 
\end{subfigure}

\caption{Retrieval results (Recall@K) for the key evaluation tasks of PinText, Related Pins, and Search.}
\label{fig:retrieval_tables}
\end{figure}

\subsection{Main Results}
The main results are presented in Figure~\ref{fig:retrieval_tables}. We find that the PinCLIP Fusion model significantly outperforms the PinCLIP image-text model on the surface retrieval tasks (+34\% Search, +44\% Related Pins) due to the introduction of (1) the cross-modal fusion architecture and (2) the neighbor alignment objective. Our method significantly outperforms prior internal content-only embedding models, such as OmniSearchSage~\cite{agarwal2024omnisearchsage} and Unified Embedding~\cite{zhai2019learning}, and state-of-the-art public multimodal embedding models~\cite{radford2021learning, zhai2023sigmoid, bolya2025perception, tschannen2025siglip, li2026qwen3}.

\subsection{Ablation Studies}
For these experiments, we train with a shorter training schedule consisting of 30k steps ($\approx$1B examples seen) using the PinCLIP Fusion model architecture, unless otherwise stated. 

\subsubsection{Impact of Model Size}
To understand the impact of model size on the task performance, we trained models at four scales.

We train with two different in-house image encoders, each of which is pretrained on a multi-label classification task~\cite{beal2022billion}. ``Hybrid-ViT-B'' (130M) is a smaller image encoder consisting of a ResNet-9~\cite{he2016deep} stem and a ``Funnel-ViT-Base-6x3'' trunk. Specifically, the Vision Transformer  consists of 3 modules with 2 intermediate funneling operations. The hidden dimension is 768, the MLP dimension is 3072, the layers per module is 6, and the number of heads is 12. ``Hybrid-ViT-g'' (922M) is a larger image encoder consisting of an InceptionNeXt-Tiny~\cite{yu2024inceptionnext} stem and a ``Funnel-ViT-giant-12x3'' trunk. In this case, the hidden dimension is 1408, the MLP dimension is 6144, the layers per module is 12, and the number of heads is 16. 

We train with two different multilingual pretrained text encoders from SigLIP~\cite{zhai2023sigmoid}. ``Transformer-B'' (278M) is a smaller text encoder consisting of 12 layers with width of 768 and 12 heads. ``Transformer-L'' (700M) is a larger text encoder consisting of  27 layers with width of 1152 and 16 heads. The vocabulary size is 250k for each model.
      
The results are presented in Table~\ref{tab:model_size}. We observe consistent gains from increased model capacity for each modality.


\begin{table}[t]
\centering
\captionsetup{justification=centering}
\setlength{\tabcolsep}{4pt}
{\small
\begin{tabular}{ll *3{S}[table-format=2.1]}
\toprule {Image Encoder} & {Text Encoder} & {R@1} & {R@5} & {R@10} \\
\midrule 
\multicolumn{5}{c}{Image-to-Text Retrieval} \\
\midrule 
Hybrid-ViT-B (130M) & Transformer-B (278M) & 56.2 & 71.1 & 75.7 \\
Hybrid-ViT-B (130M) & Transformer-L (700M) & 60.3 & 73.9 & 78.3 \\
Hybrid-ViT-g (922M) & Transformer-B (278M) & 67.2 & 79.3 & 82.7 \\
Hybrid-ViT-g (922M) & Transformer-L (700M) & 70.3 & 81.8 & 84.9 \\
\bottomrule 
\end{tabular} 
}
\caption{Assessment of the impact of model scaling per modality on PinText image-to-text retrieval performance.}
\label{tab:model_size} 
\vspace{-2em}
\end{table}

\subsubsection{Impact of Dataset Size}
We study the impact of dataset size, varying the dataset sampling ratio from 3\% to 100\% and observing the impact on image-text retrieval performance. In these experiments, the number of samples seen is held constant, with 
approximately 1B samples seen in these experiments. The results are presented in Figure~\ref{fig:datase_scaling}, showing a clear benefit to larger datasets.

\definecolor{tabblue}{RGB}{31,119,180}
\definecolor{tabred}{RGB}{214,39,40}
\definecolor{tabgreen}{RGB}{44,160,44}

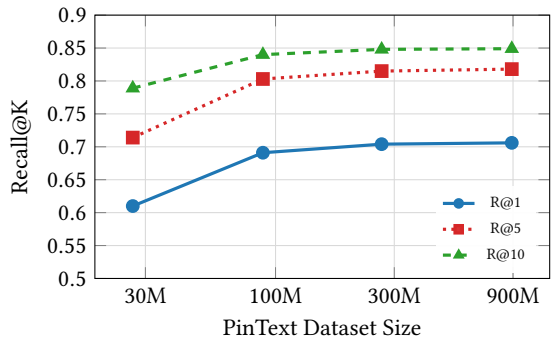
\begin{figure}[t]
  \centering
  \begin{tikzpicture}
    \begin{axis}[
      width=0.9\linewidth,
      height=0.6\linewidth,
      xlabel={PinText Dataset Size},
      ylabel={Recall@K},
      ymin=0.5, ymax=0.9,
      xtick={30,100,300,900},
      xticklabels={30M, 100M, 300M, 900M},
      xmode=log, log basis x=10,
      ytick distance=0.05,
      grid=both,
      minor grid style={gray!20},
      major grid style={gray!30},
      legend style={draw=none, fill=white, fill opacity=0.95, font=\scriptsize, inner sep=2pt, row sep=0.5pt, column sep=2pt, at={(0.97,0.02)}, anchor=south east},
      legend image post style={xscale=0.7, yscale=0.7},
      tick label style={/pgf/number format/fixed},
      every axis plot/.append style={line width=1.2pt, mark=*,
        mark options={scale=0.9, solid, draw=white, line width=0.3pt}}
    ]

    \addplot+[tabblue, solid, mark=*, mark options={solid, fill=tabblue, draw=tabblue}]
      coordinates {(26.695588,0.61) (88.975288,0.691) (266.933326,0.704) (889.767355,0.706)};

    \addplot+[tabred, dotted, mark=square*, mark options={solid, fill=tabred, draw=tabred}]
      coordinates {(26.695588,0.714) (88.975288,0.803) (266.933326,0.815) (889.767355,0.818)};
  
    \addplot+[tabgreen, dashed, mark=triangle*, mark options={solid, fill=tabgreen, draw=tabgreen}]
      coordinates {(26.695588,0.789) (88.975288,0.840) (266.933326,0.848) (889.767355,0.849)};
      
    \legend{R@1, R@5, R@10}
    \end{axis}
  \end{tikzpicture}
  \caption{Assessment of the impact of dataset scaling on PinText image-to-text retrieval performance.}
  \label{fig:datase_scaling}
\end{figure}

\subsubsection{Impact of Freezing Layers}
\begin{figure}[t]
  \centering
  \begin{tikzpicture}
    \begin{axis}[
      width=0.9\linewidth,
      height=0.6\linewidth,
      xlabel={Frozen Vision Transformer Depth},
      ylabel={Recall@K},
      ymin=0.5, ymax=0.9,
      xtick={0, 0.33, 0.67, 1.0},
      xticklabels={0\%, 33\%, 67\%, 100\%},
      ytick distance=0.05,
      grid=both,
      minor grid style={gray!20},
      major grid style={gray!30},
      legend style={draw=none, fill=white, fill opacity=0.95, font=\scriptsize, inner sep=2pt, row sep=0.5pt, column sep=2pt, at={(0.97,0.02)}, anchor=south east},
      legend image post style={xscale=0.7, yscale=0.7},
      tick label style={/pgf/number format/fixed},
      every axis plot/.append style={line width=1.2pt, mark=*,
        mark options={scale=0.9, solid, draw=white, line width=0.3pt}}
    ]

    \addplot+[tabblue, solid, mark=*, mark options={solid, fill=tabblue, draw=tabblue}]
      coordinates {(0.0,0.687) (0.33,0.703) (0.67,0.706) (1.0,0.684)};

    \addplot+[tabred, dotted, mark=square*, mark options={solid, fill=tabred, draw=tabred}]
      coordinates {(0.0,0.804) (0.33,0.814) (0.67,0.818) (1.0,0.802)};

    \addplot+[tabgreen, dashed, mark=triangle*, mark options={solid, fill=tabgreen, draw=tabgreen}]
      coordinates {(0.0,0.836) (0.33,0.847) (0.67,0.849) (1.0,0.834)};
      
    \legend{R@1, R@5, R@10}
    \end{axis}
  \end{tikzpicture}
  \caption{Assessment of the impact of freezing image encoder layers on PinText image-to-text retrieval performance.}
  \label{fig:freezing_layers}
\end{figure}
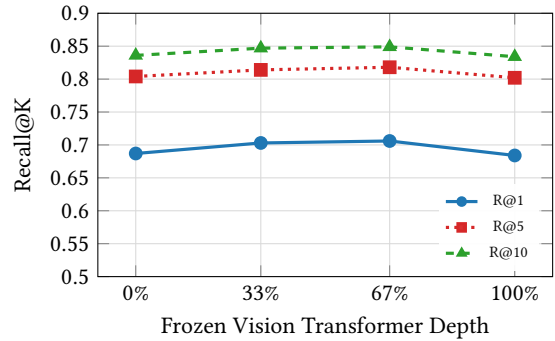

In the contrastive tuning paradigm explored by LiT~\cite{zhai2022lit}, a strong, pretrained image encoder is used as the image tower. Over the course of training, the image tower and text tower weights may be locked or unlocked. While LiT~\cite{zhai2022lit} proposes locking all of the image tower weights, we find that unlocking several of the final layers yields better performance. We study the optimal number of image encoder layers to unlock in Figure~\ref{fig:freezing_layers}. We find that unlocking the final 12 layers of the image encoder yields the best downstream performance while also significantly improving the training efficiency. Consistent with the findings in LiT~\cite{zhai2022lit}, we saw no benefit to locking layers of the text tower. In Table~\ref{tab:freezing_layers}, we report the impact on GPU memory usage and training throughput for a single-node training benchmark setup. 

\subsection{Online A/B Testing}
To evaluate the utility of our proposed multi-modal embeddings, we perform online A/B experiments across the major surfaces at Pinterest, including Homefeed, Related Pins and Search. 

\begin{table}[t]
\centering
\captionsetup{justification=centering}
\setlength{\tabcolsep}{6pt}
\begin{tabular}{S[table-format=2.0] S[table-format=2.2] S[table-format=4.0]}
\toprule
{Locked Layers} & {GPU Memory (GB)} & {Throughput (img/s)} \\
\midrule
0  & 69.37 & 1223 \\
12 & 58.19\makebox[0pt][l]{\hspace{0.3em}{\scriptsize\color{gray}(-16.1\%)}} & 1418 \makebox[0pt][l]{\hspace{0.3em}{\scriptsize\color{gray}(+15.9\%)}} \\
24 & 48.67\makebox[0pt][l]{\hspace{0.3em}{\scriptsize\color{gray}(-29.8\%)}}  & 1572\makebox[0pt][l]{\hspace{0.3em}{\scriptsize\color{gray}(+28.5\%)}} \\
36 & 41.38\makebox[0pt][l]{\hspace{0.3em}{\scriptsize\color{gray}(-40.3\%)}}  & 1658\makebox[0pt][l]{\hspace{0.3em}{\scriptsize\color{gray}(+35.5\%)}} \\
\bottomrule
\end{tabular}
\caption{Assessment of the impact of freezing image encoder layers on 1-node training efficiency (memory/throughput).}
\label{tab:freezing_layers}
\vspace{-2em}
\end{table}

\begin{table}[t]
\centering
\captionsetup{justification=centering}
\setlength{\tabcolsep}{6pt}
\begin{tabular}{lccc}
\toprule
Online Metric & Homefeed & Related Pins & Search \\
\midrule
Surface Repins & +\textbf{0.91\%} & +\textbf{1.84\%} & +\textbf{0.96\%} \\
Surface Fresh Repins & +\textbf{5.02\%} & +\textbf{14.15\%} & +\textbf{15.35\%} \\
\bottomrule
\end{tabular}
\caption{Online A/B tests by adding PinCLIP as a feature into the ranking models in Homefeed, Related Pins and Search.}
\label{tab:ranking_online}
\vspace{-1em}
\end{table}

\begin{table}[t]
\centering
\captionsetup{justification=centering}
\setlength{\tabcolsep}{6pt}
\begin{tabular}{lc}
\toprule
Online Metric & Improvement \\
\midrule
Click-Through Rate (CTR) & +\textbf{5.02\%} \\
Cost Per Click (CPC) & -\textbf{0.59\%} \\
New Ads (14d) Click Volume & +\textbf{8.67\%} \\
\bottomrule
\end{tabular}
\caption{Online A/B tests by adding PinCLIP as a feature into the Ads ranking model.}
\label{tab:ads_ranking_online}
\vspace{-2em}
\end{table}

\begin{table}[t]
\centering
\captionsetup{justification=centering}
\setlength{\tabcolsep}{6pt}
\begin{tabular}{ccc}
\toprule
\multicolumn{2}{c}{\textbf{Related Pins}} & \textbf{Search} \\
Sitewide Repins & Sitewide Fresh Repins & Fulfillment Rate \\

\midrule
+\textbf{0.36\%} & +\textbf{2.3\%} & +\textbf{0.34\%} \\
\bottomrule
\end{tabular}
\caption{Online A/B tests by using PinCLIP-based candidate generator in Related Pins and Search.}
\label{tab:retrieval_online}
\vspace{-2em}
\end{table}

\begin{figure*}[htbp] 
    \centering
    \includegraphics[width=0.7\textwidth]{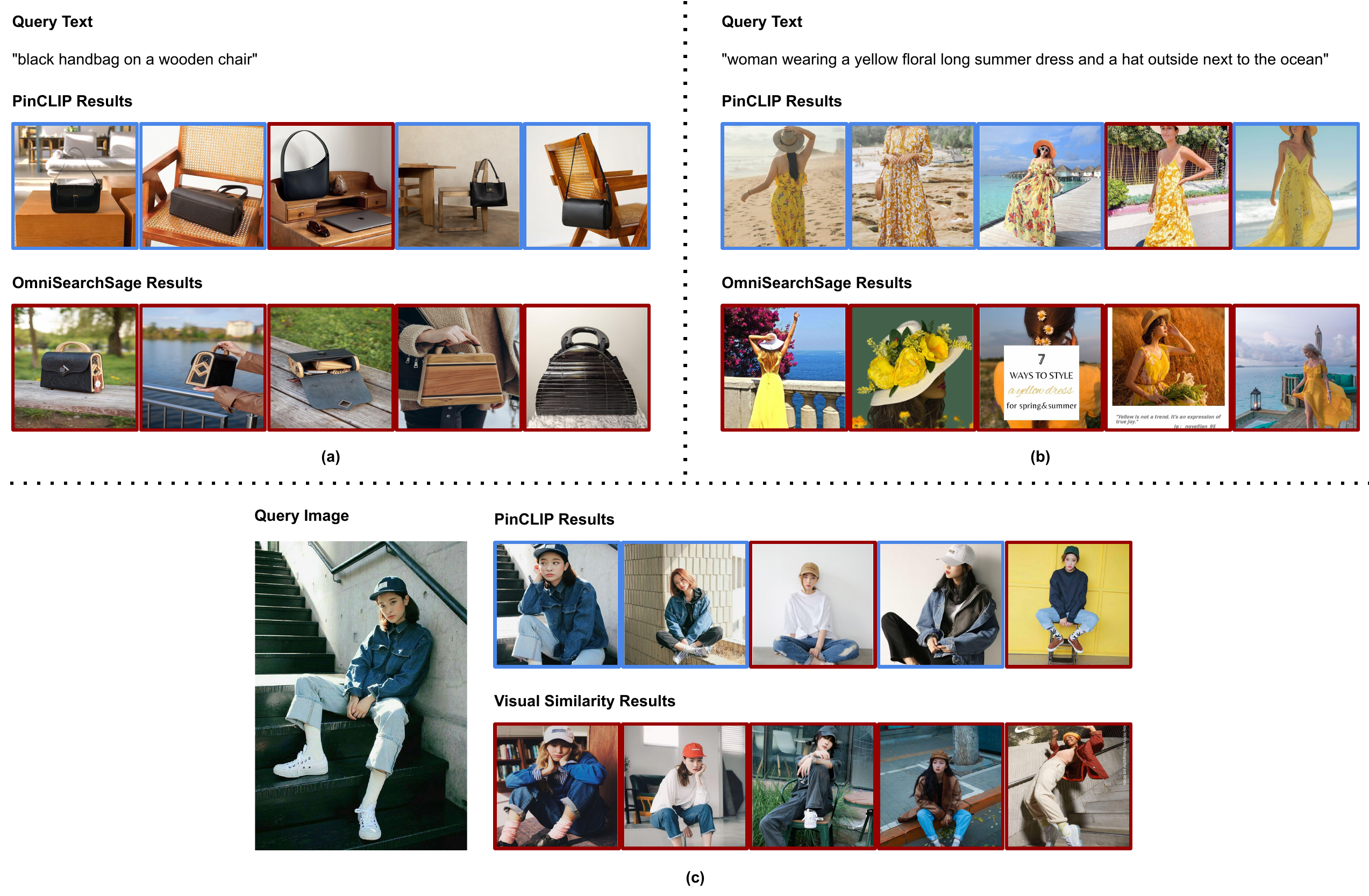} 
    \caption{Qualitative comparison of PinCLIP and OmniSearchSage~\cite{agarwal2024omnisearchsage} retrieved candidates on production-scale search corpus. Relevant candidates are highlighted in blue, whereas irrelevant candidates are highlighted in red. PinCLIP produces more semantically similar results compared to competitive baselines for both using text queries or image as inputs.}
    \label{fig:qualitative-analysis-text-and-image}
\end{figure*}

\subsubsection{Organic Ranking Models.} We added PinCLIP as a new feature into the organic ranking models, which rank non-Ads (aka. organic) content in different surfaces. We ran A/B online experiments over more than 2 weeks of data to collect the online impact. The control group is the production ranking model without the PinCLIP feature, and the treatment group is the production model with PinCLIP. For the metrics, we measure surface-level Repins, which is the most common user action type at Pinterest. Meanwhile, we also measure the number of user Repins on fresh content on each surface, where fresh content is defined as new content with age less than 28 days.

The online results are shown in Table~\ref{tab:ranking_online}. We see significant gains in surface Repins for all three main surfaces on Pinterest, which also led to significant site-wide Repin gains on Homefeed and Related Pins. The sitewide Repins for Search is not significant as the surface contributes fewer Repins, relative to the other surfaces. We also see PinCLIP is extremely effective on distributing fresh content, leading to about 15\% fresh Repin gains on Related Pins and Search.

\subsubsection{Ads Ranking Models.} We also added PinCLIP as feature into our Ads ranking models, as shown in Table~\ref{tab:ads_ranking_online}. 
We see a significant online impact in a single experiment, including a +5.02\% increase on Ads click-through rate (CTR) and a 0.59\% reduction on cost per click (CPC) for advertisers.
Consistent with the organic ranking models, we see significant improvement on fresh metrics -- an increase of 8.67\% on click volume for new Ads with age less than 14 days.

\subsubsection{Candidate Generators.} We study how to leverage PinCLIP as a new candidate generator to retrieve candidates on Search and Related Pins. On Search, we used the text embedding produced by the PinCLIP model to encode the search query, and used the fusion embedding from the PinCLIP model to encode the image corpus. On Related Pins, we used the fusion embedding to encode both the query Pin and the image corpus. For all candidate generator experiments, we used the approximate nearest neighbor (HNSW)~\cite{malkov2018efficient} algorithm to retrieve the top $k$ candidates based on the similarities of the dot product between the candidate embedding and the context embedding (search query or query Pin).

As we can see in Table~\ref{tab:retrieval_online}, the addition of the PinCLIP candidate generator in Related Pins led to a significant improvement in the number of Repins throughout the site and the number of fresh Repins throughout the site. In Search, the fulfillment rate is defined as the percentage of search sessions with positive user engagement actions, such as Repins, shares, long-clicks, etc. We also observed a significant gain in the search fulfillment rate. These experiments have demonstrated the general effectiveness of PinCLIP on both early-stage retrieval and late-stage ranking models.


\subsection{Qualitative Study}

In Figure~\ref{fig:qualitative-analysis-text-and-image}, we compare the retrieved results between the OmniSearchSage~\cite{agarwal2024omnisearchsage} and our PinCLIP fusion model, when using both text queries or image as input. As we can see, in Figure ~\ref{fig:qualitative-analysis-text-and-image}(a), PinCLIP successfully preserves object attributes (color, material) along with object-level relationships (``bag \emph{on} a chair''). In (b) PinCLIP successfully handles long text queries with multiple details. In (c), we compare PinCLIP results against our existing image embedding retrieval system. While both embeddings perform well on visual queries, we generally find that PinCLIP can yield better results. These qualitative studies further validate the general effectiveness of our proposed approach. \\

\section{Conclusion}
This paper introduces PinCLIP, a large-scale visual representation learning approach developed by Pinterest to enhance its retrieval and ranking recommendation models. We introduce a new hybrid Vision Transformer that leverages a VLM backbone and a specialized fusion and alignment mechanism to extract multi-modal representations at multiple scales. Experimental results indicate that PinCLIP exceeds the performance of leading models like Qwen by 20\% in retrieval benchmarks. 
Furthermore, the real-world impact is substantial: online testing showed widespread engagement growth across all major surfaces at Pinterest. 
Additionally, PinCLIP excels at surfacing fresh content—solving the "cold-start" issue with a 15\% jump in organic Repins and an 8.7\% increase in clicks for newly launched Ads.
\bibliographystyle{ACM-Reference-Format}
\bibliography{reference}

\appendix









\end{document}